\documentclass[letterpaper]{article} 
\usepackage{aaai2026}  
\usepackage{times}  
\usepackage{helvet}  
\usepackage{courier}  
\usepackage[hyphens]{url}  
\usepackage{graphicx} 
\usepackage{natbib}  
\usepackage{caption} 
\usepackage{algorithm}
\usepackage{algorithmic}
\usepackage{soul}
\usepackage{subcaption}  
\usepackage{newfloat}
\usepackage{listings}
\usepackage{amsfonts}
\usepackage{amsmath}
\usepackage{mathtools}
\usepackage{tabularx}
\usepackage{booktabs}   
\usepackage{multirow}   
\usepackage{graphicx}   

\DeclareCaptionStyle{ruled}{labelfont=normalfont,labelsep=colon,strut=off} 
\floatstyle{ruled}
\newfloat{listing}{tb}{lst}{}
\floatname{listing}{Listing}
\definecolor{mblue}      {RGB}{  0,114,189} 
\definecolor{mredorange} {RGB}{217, 83, 25}
\definecolor{myellow}    {RGB}{237,177, 32} 
\definecolor{mgreen}{HTML}{00AA00}
\definecolor{mblue1}{HTML}{0000FF}
\definecolor{mred}{HTML}{D40000}
\definecolor{baseline3}{HTML}{FF8080}

\graphicspath{{Figures/}}

\title{Z-Merge: Multi-Agent Reinforcement Learning for On-Ramp Merging with Zone-Specific V2X Traffic Information}
\author {
    Yassine Ibork\textsuperscript{\rm 1},
    Myounggyu Won\textsuperscript{\rm 2}, 
    Lokesh Das\textsuperscript{\rm 1}
}
\affiliations {
    \textsuperscript{\rm 1}School of Computing, Wichita State University \\
    \textsuperscript{\rm 2}Department of Computer Science, University of Memphis\\
    yxibork@shockers.wichita.edu, mwon@memphis.edu, lokesh.das@wichita.edu
}

\usepackage{bibentry}

\begin{document}

\maketitle

\begin{abstract}
Ramp merging is a critical and challenging task for autonomous vehicles (AVs), particularly in mixed traffic environments with human-driven vehicles (HVs). Existing approaches typically rely on either lane-changing or inter-vehicle gap creation strategies based solely on local or neighboring information, often leading to suboptimal performance in terms of safety and traffic efficiency. In this paper, we present a V2X (vehicle-to-everything communication)-assisted Multiagent Reinforcement Learning (MARL) framework for on-ramp merging that effectively coordinates the complex interplay between lane-changing and inter-vehicle gap adaptation strategies by utilizing zone-specific global information available from a roadside unit (RSU). The merging control problem is formulated as a Multiagent Partially Observable Markov Decision Process (MA-POMDP), where agents leverage both local and global observations through V2X communication. To support both discrete and continuous control decisions, we design a hybrid action space and adopt a parameterized deep Q-learning approach. Extensive simulations, integrating the SUMO traffic simulator and the MOSAIC V2X simulator, demonstrate that our framework significantly improves merging success rate, traffic efficiency, and road safety across diverse traffic scenarios.
\end{abstract}


\section{Introduction}

Ramp merging is a critical maneuver that involves complex interactions between merging vehicles and mainline traffic. According to the U.S. Department of Transportation, approximately 300,000 car accidents occur each year in merging areas, contributing to around 50,000 fatal crashes~\cite{el2021novel}. Ramp merging is also a major contributor to traffic congestion, often caused by poor coordination between vehicles and unpredictable driving behavior~\cite{liao2021game}. Highway merging accounts for approximately 40–80\% of traffic congestion in the U.S., and about 1.7\% of traffic fatalities occur during lane change or merging maneuvers~\cite{nakka2022multi}. 

With the rapid advancement of autonomous vehicles (AVs), numerous studies have been conducted to control AV motion in merging areas, aiming to enhance traffic efficiency, safety, and driving comfort. Two primary control strategies are inter-vehicle gap adjustment and lane-change maneuvers. Gap control approaches are primarily designed for mainline vehicles to create sufficient space for smooth merging~\cite{das2021d,das2021saint,yadavalli2023rlpg,zhu2024improving}. Alternatively, lane-change strategies focus on enabling mainline vehicles to change lanes to accommodate merging vehicles, while longitudinal control is typically used only for collision avoidance~\cite{xia2024ramp,liu2024reinforcement,cai2024graph,chen2024cooperative}. More recent approaches have focused on integrating both inter-vehicle gap adjustment and lane-change strategies to simultaneously cordinate the lane-change and gap creation for efficient merging.~\cite{wang2024cooperative,hou2023cooperative,chen2023deep}. However, these existing methods rely solely on local sensor data or limited information from neighboring vehicles via standard V2X communication, missing the opportunity to leverage ``global'' and ``zone-specific'' information that could significantly enhance traffic efficiency, safety, and driver comfort.

In this paper, we present Z-Merge, a zone-based on-ramp merging control method using multi-agent reinforcement learning (MARL). The roadway segment is strategically divided into three zones: pre-merging, merging, and ramp. Summarized traffic information from each zone, collected via a roadside unit (RSU), is independently integrated into the MARL framework. This enables agents to make more informed and holistic decisions by combining inter-vehicle gap adjustment and lane-change strategies, thereby facilitating safe, efficient, and smooth on-ramp merging. To the best of our knowledge, this is the first MARL-based merging control method that incorporates both local and zone-specific traffic information. More specifically, we formulate the problem of simultaneously controlling the motion of multiple agents in a merging area—through the complex interplay of lane-changing, acceleration/deceleration, and inter-vehicle gap adaptation—using a Multi-Agent Partially Observable Markov Decision Process (MA-POMDP) framework. The state space is carefully designed, informed by transportation literature, to incorporate critical traffic parameters that capture both local information (\emph{i.e.,} data collected through onboard sensors and vehicle-to-vehicle communication) and global (zone-specific) traffic conditions—factors that significantly influence traffic efficiency, safety, and driving comfort. To enable efficient, smooth, and collision-free merging, we design a hybrid action space that integrates lane-changing, acceleration/deceleration, and inter-vehicle gap adjustment controls. A robust multi-agent decision-making algorithm, based on the Parameterized Deep Q-Network (PDQN), is developed to select optimal control policies. We conduct extensive simulations incorporating V2X communication to train and evaluate the reinforcement learning agents. The results demonstrate that the proposed V2X-assisted MARL framework effectively enhances road safety, traffic efficiency, merging success rate, and queue length under complex merging conditions compared with state-of-the-art method. The contributions of our work are summarized as follows.

\begin{itemize}
	\item We propose the first zone-based on-ramp merging control method using multi-agent reinforcement learning (MARL) incorporating RSU-collected, zone-specific traffic information from pre-merging, merging, and ramp zones.
	\item We carefully design the state space and hybrid action space to support holistic decision-making: the state space integrates both local and global traffic information, while the hybrid action space includes discrete (lane-change) and continuous (acceleration and gap control) actions.
	\item We incorporate the Parameterized Deep Q-Network (PDQN) algorithm to achieve end-to-end vehicle control with both discrete and continuous action parameters. We develop a Double Parameterized Deep Q-Network to mitigate overestimation bias and improve learning stability.
	\item We implement a simulation framework integrating the SUMO traffic simulator~\cite{lopez2018microscopic} with Eclipse MOSAIC~\cite{schrab2022modeling} to enable realistic V2X communication and conduct extensive simulations under various traffic conditions.
\end{itemize}

\section{Related Work}
\label{sec:related-work}
A substantial body of research on on-ramp merging control strategies can be broadly categorized into three types: (1) inter-vehicle gap control-based approaches, (2) lane-change-based approaches, and (3) hybrid approaches that combine both. Inter-vehicle gap control strategies primarily focus on adjusting the spacing between vehicles on the mainline to ensure a safe merging gap for ramp vehicles. Das and Won proposed a deep Q-Network-assisted dynamic ACC method that adaptively adjusts inter-vehicle gaps~\cite{das2021d}. They later developed a safety-aware intelligent ACC using dual reinforcement learning agents to dynamically tune the time-to-collision threshold and desired gaps~\cite{das2021saint}. Yadavalli \emph{et al.} designed a reinforcement learning-based approach for dynamic intra-platoon gap adaptation to support on-ramp merging~\cite{yadavalli2023rlpg}. Zhu \emph{et al.} introduced a hierarchical gap creation strategy where mainline vehicles decelerate to create space for a platoon formed by ramp vehicles~\cite{zhu2024improving}. However, these methods are limited to controlling mainline vehicles within the merging zone, often causing upstream congestion—especially in the pre-merging area. Additionally, since they do not account for lateral maneuvers such as lane changes, these approaches are less effective in dynamically allocating road space in dense or highly interactive traffic scenarios.

Another extensively used merging control strategy is the lane-change-based approach, which regulates mandatory lane changes by merging vehicles to enter the mainline and discretionary lane changes by mainline vehicles to create space for merging. Xia \emph{et al.} proposed a multi-lane on-ramp merging strategy integrating dynamic lane-changing decisions with trajectory optimization~\cite{xia2024ramp}. Liu \emph{et al.} developed a multi-step cooperative lane-change strategy for connected autonomous vehicle (CAV) platoons approaching the merging zone~\cite{liu2024reinforcement}. Cai \emph{et al.} introduced a novel merging control method that also emphasizes mainline lane changing~\cite{cai2024graph}. Shi \emph{et al.} formulated a trajectory optimization–based lane-changing policy to enhance merge safety and minimize mainline traffic disruption~\cite{shi2024collaborative}. However, these approaches typically depend on local information and lack coordination with upstream traffic, making them less effective in handling dense or rapidly changing traffic conditions. Moreover, without integration with longitudinal gap control, they may cause sudden maneuvers or unsafe cut-ins, reducing overall traffic stability.

A hybrid on-ramp merging strategy that integrates both lane-changing and inter-vehicle gap adjustment has recently attracted increasing attention due to its adaptability, robustness, and efficiency. Wang \emph{et al.} proposed a cooperative lane-changing framework that combines longitudinal headway adjustment with lateral maneuver execution~\cite{wang2024cooperative}. Hou \emph{et al.} developed a hierarchical model for cooperative merging, where the upper layer identifies suitable merging gaps through anticipatory position-searching, and the lower layer handles lateral control~\cite{hou2023cooperative}. Chen \emph{et al.} incorporated both gap creation and mainline lane changes to facilitate merging operations~\cite{chen2023deep}. However, despite this promising trend, existing hybrid approaches largely rely on localized information or heuristic coordination, and thus fall short of exploiting the full potential of V2X communication. This limitation motivates the design of Z-Merge, which strategically incorporates zone-specific global traffic information to enable agents to make more holistic and informed merging decisions. By leveraging this broader situational awareness, Z-Merge enhances traffic efficiency, safety, and driving comfort in complex on-ramp merging scenarios.

\section{Design of Z-Merge}
\label{sec:proposed-method}

In this section, we present the details of Z-Merge. We begin with an operational overview of Z-Merge, followed by the design of our MARL framework. We then describe the implementation of our multi-agent Parameterized Deep Q-Network (PDQN) algorithm.

\subsection{Overview}
\label{sec:overview}

\begin{figure}[htbp]
	\centering
	\includegraphics[width=0.99\linewidth]{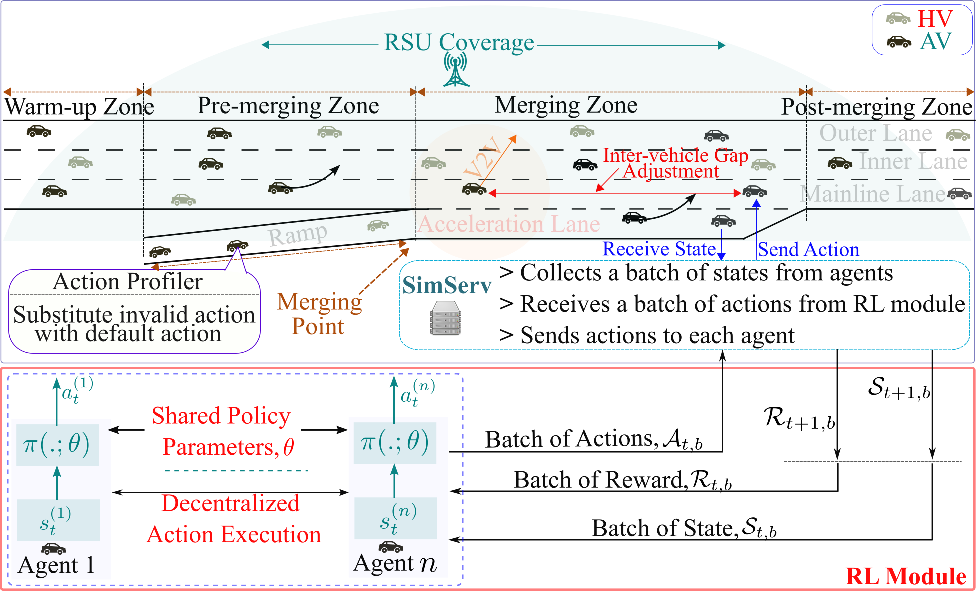}
	\caption{An operational overview of Z-Merge.}
	\label{fig:v2x-assisted-system}
\end{figure}

This section provides an operational overview of Z-Merge. The system is designed for a roadway segment with an on-ramp, as illustrated in Fig.~\ref{fig:v2x-assisted-system}. The segment is divided into three zones: the pre-merging zone, the merging zone, and the on-ramp zone. The traffic environment includes both autonomous vehicles (AVs, \emph{i.e.,} agents) and human-driven vehicles (HVs). A roadside unit (RSU) oversees the entire segment. Each vehicle gathers local information—such as position, speed, heading, and nearby vehicle states—via onboard sensors and vehicle-to-vehicle (V2V) communication, and transmits this information to the RSU through V2X communication. The RSU aggregates the received data to construct zone-specific traffic summaries, which are then made available to AVs to support coordinated and informed decision-making.

\subsection{Problem Formulation}\label{sec:problem-formulation}
We model the on-ramp merging problem using the decentralized Partially Observable Markov Decision Process (Dec-POMDP) defined by the tuple, $(\mathcal{S},\{\mathcal{A}_i,\mathcal{R}_i,\mathcal{O}_i\}_{i\in\mathcal{N}},\mathcal{T},\mathcal{O},\mathcal{Z})$. Here, $\mathcal{N}$ is the set of agents and $\mathcal{S}$ is their common state space. For each agent $i$, $\mathcal{A}_i$ is the action space, $\mathcal{R}_i:\mathcal{S}\!\times\!\mathcal{A}\!\times\!\mathcal{S}\rightarrow\mathbb{R}$ is the reward function, and $\mathcal{O}_i$ is the observation space, with joint counterparts $\mathcal{A} = \mathcal{A}_1 \times \cdots \times \mathcal{A}_n$ and $\mathcal{O} = \mathcal{O}_1 \times \cdots \times \mathcal{O}_n$. The transition kernel $\mathcal{T}:\mathcal{S}\!\times\!\mathcal{A}\!\times\!\mathcal{S}\rightarrow[0,1]$ gives the probability of moving from $s \in\mathcal{S}$ to $s' \in\mathcal{S}$ under the joint action $a\in\mathcal{A}$, while the observation kernel $\mathcal{Z}:\mathcal{S}\!\times\!\mathcal{A}\!\times\!\mathcal{O}\rightarrow[0,1]$ yields the probability of observing $o\in\mathcal{O}$ given the next state $s' \in \mathcal{S}$ and joint action $a \in \mathcal{A}$. At each step, agent $i$ receives $o_{i}\in\mathcal{O}_i$ and selects $a_{i}\in\mathcal{A}_i$, which forms part of the joint action executed in the environment. Each agent seeks a policy $\pi_i:\mathcal{O}_i\rightarrow\mathcal{A}_i$ that maximizes its expected discounted long-term reward.

\subsubsection{\textbf{State Space:}\label{state_space}} The state space of agent $i$ at time $t$, denoted by $\mathcal{S}_i^{(t)}$, incorporates both local and global traffic information. The global traffic information includes the average speed and traffic density of the pre-merging, merging, and ramp zones, as well as the queue lengths of the acceleration lane and mainline within the pre-merging zone. This information is made available to the agent via the RSU. The local information consists of both the ego vehicle’s state and the states of surrounding vehicles. The ego vehicle information—including speed, acceleration, occupied lane, and time to the merging point—is obtained through onboard sensors. The surrounding vehicle information is represented as $N_i \times F_i$, where $N_i$ is the number of vehicles observed by agent $i$, 
and $F_i$ is the feature vector for each observed vehicle, including speed, acceleration, current lane, time to the merging point (MP), and distance to the ego vehicle. This information is obtained via vehicle-to-vehicle (V2V) communication. In particular, we use the Haversine formula to compute the distances from the ego vehicle to surrounding vehicles and to the merging point, which is defined as follows.

\begin{equation}
	d= 2r\arcsin\Bigr(\sqrt{\Bigr(\sin^2\Bigr(\frac{\delta\phi}{2}\Bigr) - \cos{\phi_1\cos{\phi_2}\sin^2{\Bigr( \frac{\delta\lambda}{2}\Bigr)}} \Bigr)}  \Bigr),
	\label{eq:haversine}
\end{equation}

\noindent where $r$ is the radius of Earth, $\delta\phi = \phi_1-\phi_2$, where $\phi_1$ and $\phi_2$ are the latitudes of vehicle $1$ and vehicle $2$, $\delta\lambda = \lambda_1-\lambda_2$, where $\lambda_1$ and $\lambda_2$ are the longitudes of vehicle $1$ and vehicle $2$, respectively. Finally, the complete state space of Z-Merge at time $t$ is the product of all individual agents' state space, \emph{i.e.,} $\mathcal{S}^{(t)} = \mathcal{S}_1^{(t)} \times \mathcal{S}_2^{(t)} \times \mathcal{S}_3^{(t)}\times ...\times \mathcal{S}_{n\in \mathcal{N}}^{(t)}$.

\subsubsection{\textbf{Action Space:}} We design action spaces for both merging and mainline vehicles to enable safe, smooth, and congestion-free on-ramp merging. A hybrid action space is incorporated into our framework, allowing each RL agent to select either a discrete or continuous action. Specifically, an agent’s action space is defined as $\mathcal{A}_i = \{a_0, a_1, a_2, a_3, a_4\}$, where $a_0$ denotes a lane change to the left, $a_1$ denotes a lane change to the right, $a_2 \in [-4.5\,\mathrm{m/s^2}, 2.6\,\mathrm{m/s^2}]$ represents acceleration or deceleration, $a_3 \in [5\,\mathrm{m}, 20\,\mathrm{m}]$ specifies the desired inter-vehicle gap, and $a_4$ indicates maintaining the current state. This hybrid structure enables agents to effectively perform both lateral and longitudinal maneuvers under dynamic traffic conditions.

\subsubsection{\textbf{Reward Function:}\label{reward-function}} The reward function $\mathcal{R}_i$ for agent $i$ is composed of six sub-reward components: $r_e$ for traffic efficiency, $r_s$ for driving safety, $r_c$ for driver comfort, $r_q$ for queue length reduction, $r_d$ for deadlock avoidance, and $r_{l}$ for the effect of lane-changing behavior, which is defined as follows.

\begin{displaymath}
	\mathcal{R}_i^{(t)} = w_e r_e^{(t)} +w_s r_s^{(t)} +w_c r_c^{(t)} + w_q r_q^{(t)} + w_d r_d^{(t)} + w_{l} r_{l}^{(t)},\\
\end{displaymath}

\noindent where $w_e$, $w_s$, $w_c$, $w_q$, $w_d$, and $w_{l}$ are empirically chosen weighting coefficients. We apply a squashing function to each reward component before weighting, ensuring that the composite reward remains well-scaled and no single component dominates the learning process. Specifically, for each reward component $r_k$ (where $k=1,..., K$), we compute the squashed value as $\tilde{r}_k = \tanh{(r_k)}$, and the overall reward is then calculated as:

\begin{displaymath}
	\mathcal{R}_i^{(t)} = \sum_{k=1}^K w_k\tilde{r}_k,
\end{displaymath}

\noindent where, $w_k$ is the weighting coefficient for reward component $k$. 

\textbf{Efficiency Reward:} The efficiency reward $r_e^{(t)}$ consists of local efficiency $e_l^{(t)}$ and global traffic efficiency $e_g^{(t)}$, \emph{i.e.,} $r_e^{(t)} =  e_g^{(t)} + e_l^{(t)}$. The global efficiency $e_g^{(t)}$ quantifies effects of actions of agents on overall traffic flow, which is defined in terms of minimum average speed $\bar{v}_{e_m}$, and maximum average speed, $\bar{v}_{e_M}$ of the road segment, \emph{i.e.,}

\begin{displaymath}
	e_g^{(t)} = - (|\bar{v}_e - \bar{v}_{e_M}|)/(\bar{v}_{e_M} - \bar{v}_{e_m}).
\end{displaymath}

The local efficiency $e_l^{(t)}$ encourages the ego vehicle to maintain a speed close to the maximum speed limit $v_M$, defined for the road segment, while ensuring collision-free merging, \emph{i.e.,}

\begin{displaymath}
	e_l^{(t)} = -(|v'(t) - v_M|)/(v_M - v_m),
\end{displaymath}

\noindent where $v'^{(t)}$ and $v'_m$ are the current speed and the minimum speed of the RL agent, respectively.

\textbf{Safety Reward:} The safety reward function $r_s^{(t)}$ is designed to account for both lateral and longitudinal safety. More specifically, an agent $i$ is penalized at time $t$ if (i) it collides with another vehicle, (ii) its time-to-collision $TTC_i^{(t)}$, defined by (\ref{eq:time-to-collision}), with respect to the leading vehicle is within the time-to-collision threshold $TTC^*$ which is set to $1.2s$~\cite{ayres2001preferred} or (iii) its lateral distance $\mathcal{D}_i^{(t)}$, determined by (\ref{eq:haversine}), to the target lane leader and follower is less than the lateral safety threshold $\delta$ which is set to $12m$~\cite{dingus2006100}. Otherwise, it receives zero reward.
\begin{equation}
	TTC_i^{(t)} =
	\begin{dcases}
		\frac{(x_{n-1}-x_n-l_{n-1})}{(v_n-v_{n-1})}, & \text{if } v_n > v_{n-1}\\
		\infty, & \text{Otherwise},
	\end{dcases}
	\label{eq:time-to-collision}
\end{equation}
\noindent where $n-1$ is the leader vehicle, $n$ is the follower vehicle, $x$ is the longitudinal position, $v$ is the vehicle speed and $l$ is the length of the leader vehicle. Consequently, the safety reward function is defined as follows.

\begin{displaymath}
	r_s^{(t)}= 
	\begin{dcases}
		-1, & \text{if collision}\\
		-exp(-TTC_i^{(t)}), & \text{if } TTC_i^{(t)} \le TTC^*\\
		-\delta/\mathcal{D}, & \text{if }\mathcal{D} \le \delta \\
		0,              & \text{otherwise,}
	\end{dcases}
\end{displaymath}

\textbf{Queue Reward:} To prevent traffic congestion on the mainline of the pre-merging zone and in the acceleration lane, we penalize an agent if its actions contribute to increased queue lengths in these areas. The queue-based penalty is defined as
\[
r_q^{(t)} = -\log_{10}(\mathcal{Q}_p^{(t)} + \mathcal{Q}_a^{(t)}),
\]
where $\mathcal{Q}_p^{(t)}$ and $\mathcal{Q}_a^{(t)}$ represent the queue lengths on the pre-merging zone and in the acceleration lane at time $t$, respectively.

\textbf{Driving Comfort:} The driving comfort reward function \( r_c^{(t)} \), adopted from~\cite{xu2024graph}, penalizes excessive acceleration or deceleration and is defined as follows.
\begin{displaymath}
	r_c^{(t)} =
	\begin{dcases}
		-(|\alpha_i^{(t)}|-\alpha_{max})/(|\alpha_i^{(t)}|) & \text{if } |\alpha_i^{(t)}| > \alpha_{max}\\
		0 & \text{otherwise},
	\end{dcases}
\end{displaymath}
where \( |\alpha_i^{(t)}| \) denotes the magnitude of the longitudinal acceleration (or deceleration) applied by agent \( i \) at time \( t \), and \( a_{\text{max}} \) is the comfort threshold, set to \( 2.6\,\mathrm{m/s^2} \).

\textbf{Deadlock Penalty:} To prevent merging vehicles from remaining stagnant at the end of the acceleration lane, we define the deadlock reward $r_d^{(t)}$ to penalize agents more heavily as they approach the end of the lane, \emph{i.e.,}

\begin{displaymath}
	r_d^{(t)}=
	\begin{dcases}
		- exp\Bigr(-\frac{(x-L)^2}{10L}\Bigr), & \text{if stay on accel. lane}\\
		0, & \text{otherwise},
	\end{dcases}
\end{displaymath}

\noindent where $x$ is the current position of the ego vehicle on the acceleration lane, and $L$  is the length of the acceleration lane.

\textbf{Lane Change Penalty:} Excessive lane changes pose challenges to traffic safety and reduce traffic efficiency. We incorporate a lane change reward function $r_l^{(t)}$ to penalize an agent if it performs unnecessary lane changes, which is defined as follows.

\begin{displaymath}
	r_{l}^{(t)} =
	\begin{dcases}
		-1, & \text{if  change lanes} \\
		0 , & \text{otherwise}.
	\end{dcases}
\end{displaymath}

\section{Multi-Agent Parameterized Double DQN}
\label{sec:mapddqn}

Due to the hybrid nature of our action space, we solve our multi-agent on-ramp merging problem using the Parameterized Deep Q-Network (PDQN) algorithm~\cite{xiong2018parametrized}. In particular, we adopt the centralized training with decentralized execution (CTDE) paradigm~\cite{amato2024introduction}, along with a parameter-sharing mechanism: all agents share a common policy and value network and contribute to a shared replay buffer by recording their experiences at each time step. This approach enables each agent to benefit from a larger and more diverse dataset, which accelerates learning while maintaining decentralized decision-making during execution.

In a hybrid Markov Decision Process (MDP), the action space is defined as a joint space
\[
\mathcal{A} = \left\{(k, x_k)\;\middle|\;k \in \{1, \dots, K\},\; x_k \in \mathcal{X}_k \right\},
\]
where $k$ is a discrete high-level action index and $x_k$ is the corresponding continuous argument. The policy must select both a discrete action and its associated continuous argument. Following~\cite{xiong2018parametrized}, we employ an actor–critic architecture: a Q-network $Q_{\phi}(s, k, x_k)$ evaluates state–action–parameter tuples, and a parameter (actor) network $\mu_{\theta}(s) \in \mathbb{R}^{\sum_k d_k}$ outputs a joint vector $x = (x_1, \dots, x_K)$, where each $x_k$ corresponds to the continuous argument for discrete action $k$.

At each time step $t$, the parameter network generates a full vector of continuous parameters $x_t = \mu_{\theta}(s_t)$. The Q-network then computes the Q-values for each discrete action using its corresponding parameter vector component, and selects the action as $k^\ast = \arg\max_{k} Q_{\phi}(s_t, k, x_k)$.

To address overestimation bias, we adopt a Double PDQN approach. For each transition $(s_t, a_t, r_t, s_{t+1})$, we first select the candidate action using the online critic:
\[
k^{\text{online}\,*} = \arg\max_{k'} Q_{\phi}\left(s_{t+1}, k', \mu_{\theta}(s_{t+1})_{k'}\right),
\]
then evaluate it with the target critic:
\begin{equation}
	y_t = r_t + \gamma\, Q_{\phi^-}\left(s_{t+1}, k^{\text{online}\,*}, \mu_{\theta^-}(s_{t+1})_{k^{\text{online}\,*}}\right),
	\label{eq:doubleq_target_pdqn}
\end{equation}
where $\phi^-$ and $\theta^-$ denote the parameters of the target Q-network and target parameter network, respectively. These target parameters are updated periodically using hard updates every 35,000 gradient steps.

The Q-network is optimized by minimizing the temporal-difference (TD) error:
\[
\mathcal{L}_{Q} = \left(Q_{\phi}(s_t, k_t, x_{k_t}) - y_t\right)^2.
\]
Meanwhile, the parameter network is updated to maximize the expected Q-value by minimizing the negative of the estimated return:
\[
\mathcal{L}_{\mu} = -\,\mathbb{E}_{s \sim \mathcal{D}} \left[ Q_{\phi}\left(s, k, \mu_{\theta}(s)_k\right) \right].
\]
The hyperparameters used in the optimization process are summarized in Table~\ref{tab:simulation_parameters}.

\section{Simulation Results}
\label{sec:simulation-results}
\subsection{Simulation Setup\label{sumilation-setup}}

We employed traffic simulators MOSAIC~\cite{schrab2022modeling} and SUMO~\cite{lopez2018microscopic} to implement Z-Merge. SUMO is used to simulate the microscopic vehicle dynamics and road traffic environment, whereas MOSAIC is employed to ensure V2X communication. The MARL framework of Z-Merge was developed using PyTorch. To interface the framework with MOSAIC—which is implemented in Java—we developed a communication module called ``SimServ''. SimServ acts as a bridge between the simulation environment and the RL module: it aggregates the state information of each RL agent and transmits it to the RL framework. The RL module then processes a batch of states and returns a corresponding batch of actions to SimServ. SimServ forwards each action to the appropriate agent using predefined routing identifiers. Upon receiving an action, each agent invokes an action validation module (``action profiler'') to determine whether the action is valid. If the action is deemed invalid, it is replaced with a default fallback action.

We consider a $320\,\mathrm{m}$-long road segment with an on-ramp, which is divided into a $150\,\mathrm{m}$ pre-merging zone and a $100\,\mathrm{m}$ merging zone. The on-ramp is $100\,\mathrm{m}$ in length. Vehicles use initial $50\,\mathrm{m}$ called "warm-up zone" for injecting into the network and adapting the speed limit. Traffic flow rates are set to $3600$ and $900$ vehicles per hour per lane for the mainline and the ramp, respectively, with vehicle arrivals modeled as a Poisson process. Among all vehicles, $60\%$ are AVs controlled by Z-Merge, while the remaining $40\%$ are HVs governed by MOSAIC's default driving model. Each AV agent makes decisions regarding throttle, braking, and lane changes at intervals of $0.1\,\mathrm{s}$.

We performed hyperparameter tuning to identify the optimal configuration for achieving the best performance. Specifically, we used a grid search method to select the optimal values for the learning rate, discount factor, batch size, replay buffer size, target update frequency, and the number of hidden layers in both the actor and parameter networks. The final hyperparameters used in the simulation are summarized in Table~\ref{tab:simulation_parameters}.

We validated Z-Merge using five key evaluation metrics: (i) traffic efficiency, (ii) traffic safety, (iii) driving comfort, (iv) success rate, and (v) queue length. We compared our method against three baseline approaches: ``lane change'' (hereafter, Baseline-1)~\cite{liu2024reinforcement}, ``inter-vehicle gap adjustment'' (Baseline-2)~\cite{zhu2024improving}, and ``lane change with inter-vehicle gap adjustment in the merging zone'' (Baseline-3)~\cite{wu2025ppo}. Traffic efficiency is measured by the average speed of all vehicles on the highway segment. Traffic safety is quantified by the collision rate, defined as the proportion of AVs that collide with other vehicles during the simulation. Driving comfort reflects the smoothness of driving and is evaluated based on vehicle acceleration and deceleration during the simulation~\cite{de2023standards}. The success rate is defined as the fraction of vehicles entering from the ramp that successfully merge into the mainline without collisions or coming to a stop. Queue length is defined as the number of consecutive vehicles traveling at speeds below $2\,\mathrm{m/s}$~\cite{li2025multi}.

\begin{table}[htbp]
	\centering
	\caption{Hyper/parameters for Z-Merge.}
	\label{tab:simulation_parameters}
	\begin{tabular}{ll}
		\hline
		\multicolumn{2}{c}{\textbf{Environment Parameters}}\\
		\hline
        V2V \& V2X range & $(100m, 500m)$ \\
		Highway speed limit, $v_{max}$ & $32m/s$\\ 
        Accel. range, $(a_{max}, a_{min})$ & $a \in [2.8, -4.5]m/s^2$\\ 
		Simulation time step, $\Delta t$ & $0.1s$\\
		Time gap offset, $\tau$ & $1s$\\
		\multicolumn{2}{c}{\textbf{RL Parameters}}\\
		\hline
		State space, action space & $42,\;5$ \\
		Hidden layers             & $[256,\,512,\,512,\,128]$ \\
		Activation function       & ReLU\\
		Loss function             & Huber Loss \\
		$(\epsilon_{\mathrm{init}},\,\epsilon_{\mathrm{final}},\,\epsilon_{\mathrm{decay}})$ 
		& $(1.0,\,0.01,\,0.999985)$ \\
		Batch size, $(b)$          & $64$ \\
		Discount factor, $(\gamma)$& $0.995$ \\
		Replay buffer size, $\mathcal{B}$ & $100{,}000$ \\
		Learning rate\\(Actor, Actor-Param)                  & ($1\times10^{-4}$, $1\times10^{-4}$)  \\
		\hline
	\end{tabular}
	\vspace{-10pt}
\end{table}

\subsection{Traffic Efficiency}

\begin{figure}[ht!]
	\centering
    \includegraphics[width=\linewidth]{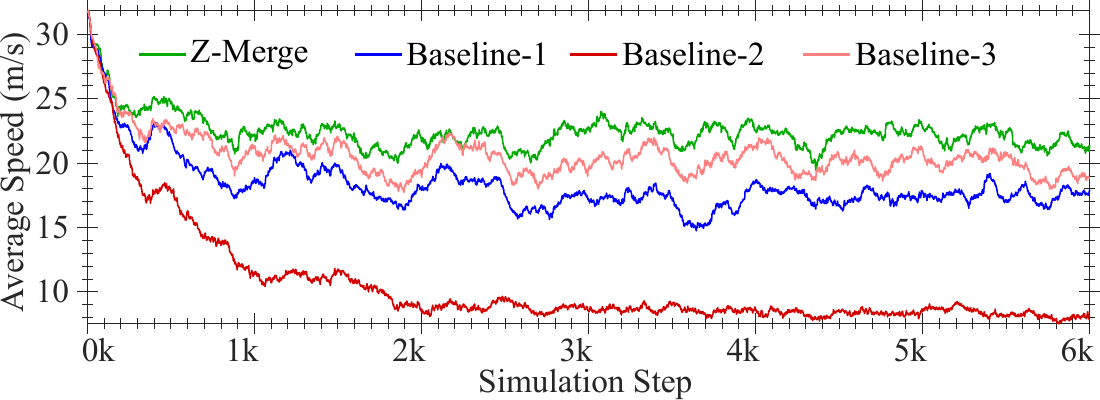}
	\caption{The average speed changes over time.}
	\label{fig:avg-speed-over-sim-step}
\end{figure}

\begin{figure}[ht!]
	\centering
	\includegraphics[width=0.99\linewidth, ]{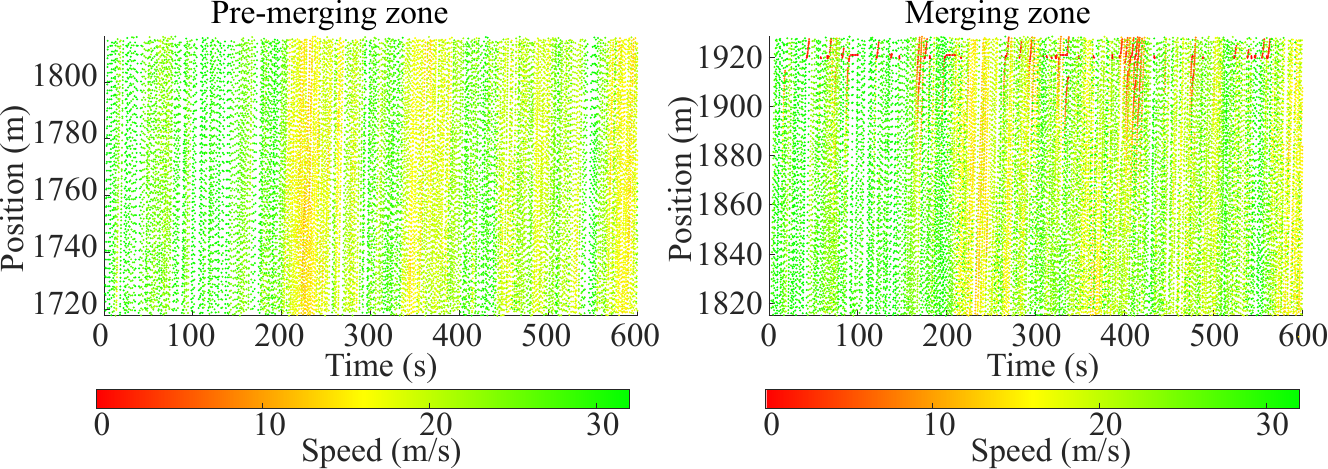}
	\caption{Spatio-temporal representation of the average speed for Z-Merge in pre-merging and merging zones.}
	\label{fig:spatio-temporal}
\end{figure}

We measured traffic efficiency in terms of the average vehicle speed. Fig.~\ref{fig:avg-speed-over-sim-step} presents the average speed over time for all methods. Initially, the differences among the methods were minimal, as vehicles entered the simulation at the maximum initial speed. However, as traffic density increased, the advantages of coordinated lane-changing and inter-vehicle gap adjustment became more apparent. In particular, Z-Merge consistently achieved $26.05\%$ and $131.34\%$ higher average speed compared to Baseline-1 and Baseline-2, respectively, throughout the simulation. This improvement is attributed to Z-Merge’s ability to reduce stop-and-go traffic in the pre-merging and merging zones, as illustrated in the space–time diagram for Z-Merge (Fig.~\ref{fig:spatio-temporal}), which highlights its effective management of the complex interplay between lane changes and inter-vehicle gap adjustments. In comparison with Baseline-3, Z-Merge achieves slightly better performance by $10\%$. While the gain is modest, it is still noteworthy given that Z-Merge maintains higher traffic efficiency and also delivers consistently strong results across other evaluation metrics, as discussed in the following sections.

We also assess the impact of agent penetration rates (PRs) on traffic efficiency, where PRs represent the proportion of AVs relative to HVs in the simulation. Table~\ref{tab:performance-over-pr} summarizes the results. Compared to Baseline-1 and Baseline-2, Z-Merge yields significantly higher traffic efficiency—up to a $118.6\%$ improvement—as PRs increase from $5\%$ to $60\%$. Interestingly, PRs have minimal or even negative effects on traffic efficiency for Baseline-2. This is because, although mainline vehicles create inter-vehicle gaps, some AVs are forced to decelerate, causing congestion in the pre-merging zone. Furthermore, the absence of a lane-change module prevents merging vehicles from utilizing available gaps effectively, leading to congestion in the acceleration lane. We also observe that Baseline-3 performs similar or slightly better than Z-Merge at high PRs, but it struggles to maintain high efficiency at lower PRs. We need to change this line. Notably, this performance gain at high PRs comes at the cost of reduced safety and success rate. This suggests that relying solely on local information can yield higher efficiency but may compromise safety. In contrast, Z-Merge's panoramic control over both the merging and pre-merging zones enables it to achieve high traffic efficiency while also preserving safety and merging success (discussed in the following sections).

\begin{table*}[htbp]
	\centering
	\setlength{\tabcolsep}{2.5pt}
	\renewcommand{\arraystretch}{1.75}
	\caption{Effect of PRs on the performance of Z-Merge and Baselines.}
	\scalebox{0.67}{%
		\begin{tabular}{c*{3}{c}*{3}{c}*{3}{c}*{3}{c}}
			\toprule
			\multirow{2}{*}{PR (\%)} &
			\multicolumn{3}{c}{Z-Merge} &
			\multicolumn{3}{c}{Baseline-1} &
			\multicolumn{3}{c}{Baseline-2} &
			\multicolumn{3}{c}{Baseline-3} \\
			\cmidrule(lr){2-4}\cmidrule(lr){5-7}\cmidrule(lr){8-10}\cmidrule(lr){11-13}
			& Efficiency & Safety (\%) & Succ.\ Rate (\%) &
			Efficiency & Safety (\%) & Succ.\ Rate (\%) &
			Efficiency & Safety (\%) & Succ.\ Rate (\%) &
			Efficiency & Safety (\%) & Succ.\ Rate (\%) \\
			\midrule
			5  & $\mathbf{10.65}\pm\mathbf{0.86}$ & $6.27\pm3.79$  & $16.46\pm6.84$ &
			$10.24\pm0.75$ & $11.26\pm6.88$ & $7.92\pm5.85$ &
			$9.56\pm0.73$  & $\mathbf{0.00}\pm\mathbf{0.00}$  & $16.42\pm4.10$ &
			$9.82\pm1.02$ & $6.32\pm3.82$ & $\mathbf{17.50}\pm\mathbf{5.85}$ \\
			10 & $\mathbf{12.31}\pm\mathbf{1.42}$ & $5.31\pm3.20$  & $\mathbf{25.31}\pm\mathbf{7.63}$ &
			$11.35\pm0.87$ & $9.06\pm3.74$  & $28.19\pm5.75$ &
			$9.43\pm0.73$  & $\mathbf{0.07}\pm\mathbf{0.31}$  & $20.92\pm4.14$ &
			$9.06\pm2.01$ & $4.95\pm2.63$ & $20.82\pm5.20$ \\
			20 & $\mathbf{16.38}\pm\mathbf{1.68}$ & $4.99\pm2.01$  & $\mathbf{51.15}\pm\mathbf{11.20}$ &
			$13.24\pm1.29$ & $6.88\pm1.98$  & $43.42\pm5.93$ &
			$9.41\pm0.57$  & $\mathbf{0.10}\pm\mathbf{0.32}$  & $28.16\pm4.48$ &
			$12.15\pm3.71$ & $5.01\pm1.78$ & $38.84\pm11.10$ \\
			40 & $\mathbf{21.58}\pm\mathbf{0.91}$ & $4.26\pm1.29$  & $\mathbf{85.23}\pm\mathbf{4.99}$ &
			$17.12\pm1.82$ & $3.52\pm1.46$  & $76.51\pm7.90$ &
			$9.34\pm0.72$  & $\mathbf{0.02}\pm\mathbf{0.08}$  & $45.47\pm5.36$ &
			$19.61\pm2.99$ & $6.13\pm1.69$ & $73.61\pm6.65$ \\
			60 & $\mathbf{23.28}\pm\mathbf{0.78}$ & $2.66\pm0.88$  & $\mathbf{93.25}\pm\mathbf{2.85}$ &
			$18.10\pm1.33$ & $2.14\pm0.87$  & $89.51\pm4.75$ &
			$9.12\pm0.91$  & $\mathbf{0.07}\pm\mathbf{0.13}$  & $59.36\pm5.03$ &
			$22.38\pm1.10$ & $6.21\pm1.35$  & $82.61\pm4.68$ \\
			\bottomrule
		\end{tabular}}
	\label{tab:performance-over-pr}
	\vspace{-8pt}
\end{table*}

\subsection{Traffic Safety\label{sec:traffic-safety}}

We evaluated the performance of Z-Merge in terms of traffic safety, measured by the collision rate. The results are summarized in Table~\ref{tab:performance-over-pr}. Among all approaches, Baseline-2 achieves the best traffic safety, with its collision rate never exceeding $0.11\%$ across all PRs. This is because Baseline-2 only adjusts inter-vehicle gaps and does not perform lane changes, thereby eliminating additional collision risks introduced by lateral maneuvers. In contrast, Baseline-3 exhibits significantly higher collision rates, particularly at low PRs. For example, at a PR of $5\%$, the collision rate reaches $21.20\%$. However, the collision rate for Baseline-3 decreases as PR increases, dropping to $4.53\%$ at $60\%$ PR. This trend suggests that a higher presence of AVs mitigates the risk of unsafe interactions, even without global coordination. Z-Merge significantly reduces the collision rate compared to Baseline-3—by up to $57.16\%$—while achieving safety levels comparable to Baseline-1. Moreover, Z-Merge’s collision rate steadily declines as PR increases, reaching $2.66\%$ at $60\%$ PR. This improvement can be attributed to Z-Merge’s panoramic coordination across both merging and pre-merging zones. Unlike Baseline-3, which allows full local control of lane changes and gap adjustments but lacks information from surrounding zones, Z-Merge leverages cross-zone information to proactively manage interactions and reduce conflicts.

\subsection{Driving Comfort}
\label{driving-comfort}

\begin{figure}[htbp]
	\centering
	\begin{subfigure}{0.32\columnwidth}
		\includegraphics[width=0.99\textwidth]{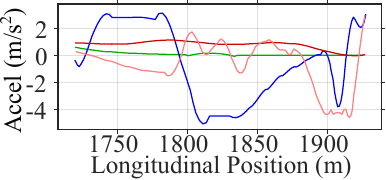}
		\caption{Accel at 200s}
		\label{fig:acc-merging-vehicle-100}
	\end{subfigure}\hfill
	\begin{subfigure}{0.32\columnwidth}
		\includegraphics[width=0.99\textwidth]{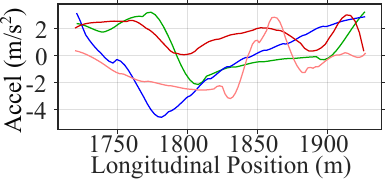}
		\caption{Accel at 290s} 
		\label{fig:acc-merging-vehicle-290}
	\end{subfigure}\hfill
	\begin{subfigure}{0.32\columnwidth}
		\includegraphics[width=0.99\textwidth]{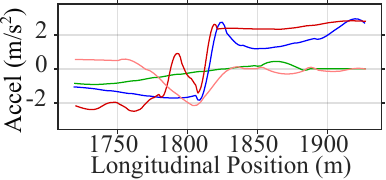}
		\caption{Accel at 360s}
		\label{fig:acc-merging-vehicle-360}
	\end{subfigure}\hfill
	\vspace{5pt}
	\begin{subfigure}{0.32\columnwidth}
		\includegraphics[width=0.99\textwidth]{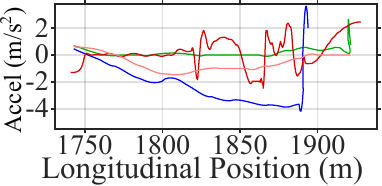}
		\caption{Accel at 200s}
		\label{fig:acc-mainline-vehicle-100}
	\end{subfigure}\hfill
	\begin{subfigure}{0.32\columnwidth}
		\includegraphics[width=0.99\textwidth]{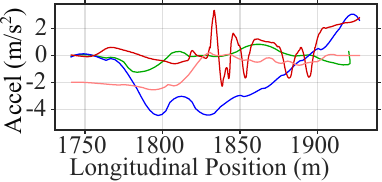}
		\caption{Accel at 290s}
		\label{fig:acc-mainline-vehicle-290}
	\end{subfigure}\hfill
	\begin{subfigure}{0.32\columnwidth}
		\includegraphics[width=0.99\textwidth]{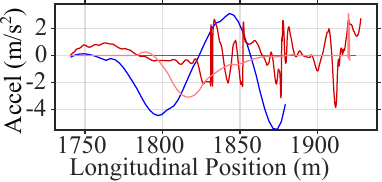}
		\caption{Accel at 360s}
		\label{fig:acc-mainline-vehicle-360}
	\end{subfigure}
	\caption{Acceleration trajectories for vehicles at three different time instances. The top row presents the acceleration profile of a merging vehicle, whereas the bottom row presents that of a mainline vehicle.
		\textcolor{mgreen}{\rule[2pt]{10pt}{1.2pt}} Z-Merge,
		\textcolor{mblue1}{\rule[2pt]{10pt}{1.2pt}} Baseline-1,
		\textcolor{mred}{\rule[2pt]{10pt}{1.2pt}} Baseline-2,
		\textcolor{baseline3}{\rule[2pt]{10pt}{1.2pt}} Baseline-3.}
	\label{fig:driving-comfort-at-different-approach}
\end{figure}

To evaluate driving comfort, we analyze the acceleration profiles of vehicles injected into the simulation at three representative time points: $200\,s$, $290\,s$, and $360\,s$. For each time point, we track one mainline vehicle and one ramp vehicle that experience the highest variation in acceleration or deceleration. The resulting profiles are shown in Fig.~\ref{fig:driving-comfort-at-different-approach}. It is evident that AVs controlled by Z-Merge exhibit smoother and more continuous acceleration patterns. In particular, mainline vehicles maintain a gentle acceleration profile, while ramp vehicles experience minimal fluctuations. In contrast, AVs using Baseline-1 and Baseline-2 show sharp acceleration changes, especially within the merging zone (\emph{i.e.,} positions between $1830–1950\,m$). Baseline-3 also struggles to maintain smooth acceleration, particularly for mainline vehicles. Notably, Z-Merge maintains acceleration within the comfortable driving range. According to~\cite{de2023standards}, a comfortable acceleration threshold is defined as $1.47\,\mathrm{m/s}^2$, and accelerations up to $2.12\,\mathrm{m/s}^2$ are considered acceptable. Z-Merge stays well within these limits across all scenarios.

\subsection{Queue Length}

\begin{figure}[h]
	\centering
	\begin{minipage}{.49\columnwidth}
		\centering
		\includegraphics[width=\linewidth]{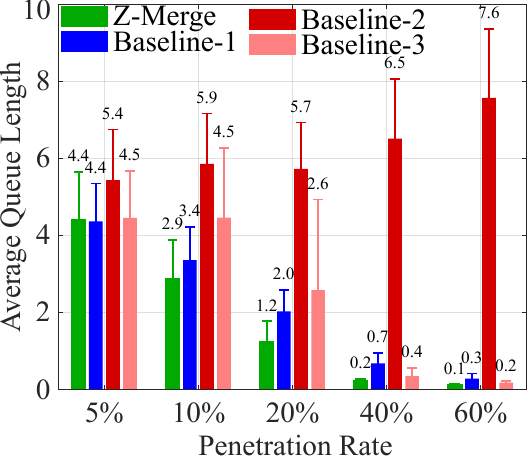}
		\caption{Avg. queue length.}
		\label{fig:queue-length}
	\end{minipage}%
	\hspace*{3mm}
	\begin{minipage}{.49\columnwidth}
		\centering
		\includegraphics[width=\linewidth]{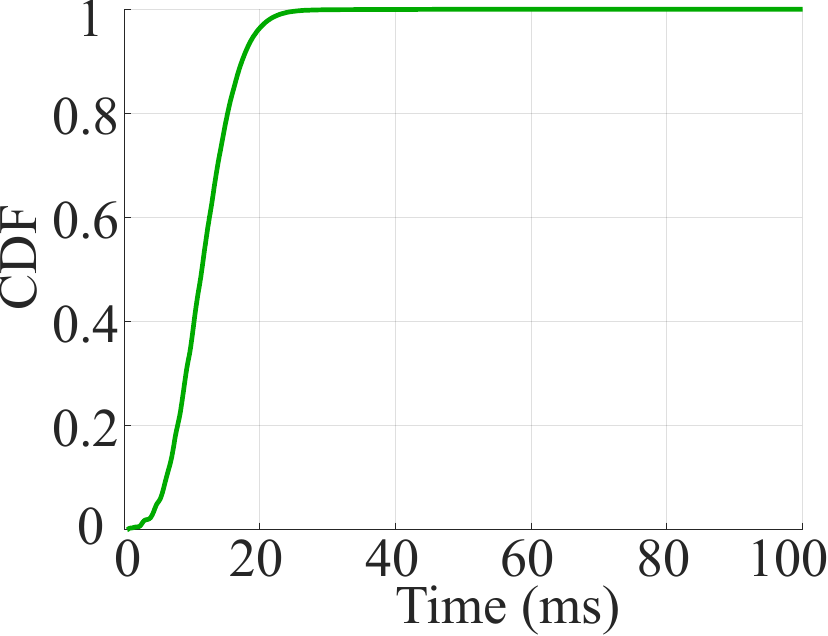}
		\caption{Computational overhead.}
		\label{fig:computational-overhead}
	\end{minipage}
\end{figure}

We measured the average queue length with respect to varying PRs. Fig.~\ref{fig:queue-length} shows the results. We observe that queue length decreases for Baseline-1, Baseline-3, and Z-Merge as PRs increase, primarily because the growing presence of AVs enables smoother merging and reduced stop-and-go behavior. Interestingly, however, the queue length for Baseline-2 increases with higher PRs. This is because Baseline-2 relies heavily on creating larger inter-vehicle gaps to facilitate merging, which causes upstream vehicles to decelerate and contributes to longer queues. Overall, Z-Merge achieves the shortest queue lengths across all PR levels, outperforming Baseline-1, Baseline-2, and Baseline-3 by up to $65.51\%$, $98.68\%$, and $50\%$, respectively. These results highlight the effectiveness of combining zone-specific global information with synchronized control of both lane-changing and inter-vehicle gap adjustment.

\subsection{Success Rate \label{sec:success-rate}}

The success rate indicates the proportion of vehicles that successfully enter the mainline without stopping at the end of the acceleration lane or colliding with other vehicles. As shown in Table~\ref{tab:performance-over-pr}, the success rate of Z-Merge increases significantly with higher PRs of AVs on both the ramp and the mainline. Specifically, Z-Merge improves the success rate from $16.4\%$ at a PR of $5\%$ to $93.25\%$ at $60\%$, representing a relative gain of $466.52\%$. Although Baseline-1, Baseline-2, and Baseline-3 also show improvement with increasing PRs, their success rates remain substantially lower than that of Z-Merge. On average, they underperform Z-Merge by $4.17\%$, $57.09\%$, and $12.87\%$, respectively. This performance gap can be attributed to their limited capabilities: Baseline-1 and Baseline-2 rely solely on either lane-changing or inter-vehicle gap adjustment, while Baseline-3 lacks access to global, zone-specific information. As a result, these baselines struggle to coordinate effectively in the dynamic and complex context of on-ramp merging.

\subsection{Computational Overhead}

Agents must make decisions rapidly on high-speed roadways. In particular, continuous control over lane changes and inter-vehicle gap adjustments in complex and dynamically evolving scenarios—such as on-ramp merging—requires minimal computational delay. In this section, we measured the inference time of the RL decision-making process. Specifically, we evaluated the per-step decision interval over 6,000 iterations on a Mac Mini equipped with an M4-Pro chip and 48\,GB of RAM. Each cycle includes (i) processing the current batch of states, (ii) computing the corresponding actions, (iii) sending actions to the environment, and (iv) receiving the next states. Fig.~\ref{fig:computational-overhead} presents the cumulative distribution function (CDF) of the per-step inference time. The results indicate that in 90\% of cases, the inference time remains below $16\,ms$, with an average delay of $11.70\,ms$. Overall, the observed computational latency is sufficiently low to support real-time decision-making in high-speed traffic environments.

\section{Conclusion}
\label{sec:conclusion}

We presented Z-Merge, a MARL framework designed to enhance the efficiency of on-ramp merging through synchronized lane-change and inter-vehicle gap adjustment strategies. By leveraging zone-specific global information and centralized training with decentralized execution, Z-Merge demonstrated superior performance across multiple metrics—including traffic efficiency, safety, driving comfort, success rate, and queue length—when compared to existing baselines. Extensive simulation results under varying penetration rates confirmed the effectiveness of Z-Merge. Furthermore, inference time analysis showed that Z-Merge operates within real-time constraints, making it suitable for high-speed driving environments. 

\bibliography{References/IEEEabrv, References/rampmerging-bib}


\end{document}